\documentclass[final]{cvpr}

\usepackage{times}
\usepackage{epsfig}
\usepackage{graphicx}
\usepackage{amsmath}
\usepackage{amssymb}

\usepackage{algorithm}  
\usepackage{algpseudocode}  
\usepackage{amsmath}  
\usepackage{adjustbox}
\usepackage{booktabs}
\usepackage{multirow}
\usepackage{enumitem}
\usepackage[pagebackref=true,breaklinks=true,colorlinks,bookmarks=false]{hyperref}
\def\ie{\textit{i.e.}}

\def\etal{\textit{et al.}}

\def\L{\mathcal{L}}
\def\M{\mathcal{M}}
\def\E{\mathit{E}}

\def\cvprPaperID{7481} 
\def\confYear{CVPR 2021}
\begin{document}

\title{Semi-supervised Domain Adaptation based on Dual-level Domain Mixing for Semantic Segmentation}

\author{Shuaijun Chen$^{1\dagger}$, Xu Jia$^{2\dagger}\footnotemark[3]$, Jianzhong He$^{1,3}$, Yongjie Shi$^{1,4}$, Jianzhuang Liu$^1$\\
$^1$Noah's Ark Lab, Huawei Technologies. $^2$Dalian University of Technology.\\
$^3$Data Storage and Intelligent Vision Technical Research Dept, Huawei Cloud.\\
$^4$Key Lab of Machine Perception, Peking University\\
{\tt\small \{chenshuaijun,jianzhong.he,shiyongjie2,liu.jianzhuang\}@huawei.com,xjia@dlut.edu.cn}
}

\maketitle

\renewcommand{\thefootnote}{\fnsymbol{footnote}}
\footnotetext[2] {Equal contribution}
\footnotetext[3]{Part of this work was done while he was in Noah's Ark Lab}

\setlength{\abovedisplayskip}{6pt}
\setlength{\belowdisplayskip}{6pt}
\setlength{\abovedisplayshortskip}{0pt}
\setlength{\belowdisplayshortskip}{0pt}

\begin{abstract}
Data-driven based approaches, in spite of great success in many tasks, have poor generalization when applied to unseen image domains, and require expensive cost of annotation especially for dense pixel prediction tasks such as semantic segmentation. Recently, both unsupervised domain adaptation (UDA) from large amounts of synthetic data and semi-supervised learning (SSL) with small set of labeled data have been studied to alleviate this issue.
However, there is still a large gap on performance compared to their supervised counterparts. We focus on a more practical setting of semi-supervised domain adaptation (SSDA) where both a small set of labeled target data and large amounts of labeled source data are available. To address the task of SSDA, a novel framework based on dual-level domain mixing is proposed. The proposed framework consists of three stages. First, two kinds of data mixing methods are proposed to reduce domain gap in both region-level and sample-level respectively. We can obtain two complementary domain-mixed teachers based on dual-level mixed data from holistic and partial views respectively. Then, a student model is learned by distilling knowledge from these two teachers. Finally, pseudo labels of unlabeled data are generated in a self-training manner for another few rounds of teachers training. Extensive experimental results have demonstrated the effectiveness of our proposed framework on synthetic-to-real semantic segmentation benchmarks.
\end{abstract}

\section{Introduction}
\label{intro}
Semantic segmentation with the goal of assigning semantic-level labels to every pixel in an image is one of the fundamental topics in computer vision due to its widely critical real-world applications, such as autonomous driving~\cite{geiger2012we} and robotic navigation~\cite{milioto2018real, shvets2018automatic}.
Over the past few years, deep convolutional neural networks(CNNs) have achieved dramatic improvements in semantic segmentation~\cite{chen2017deeplab, liu2015semantic, hoffman2016fcns, luc2017predicting, chen2018encoder, zhao2017pyramid}. The success of CNN-based methods benefits from large volume of manually labeled data~\cite{lin2014microsoft, everingham2015pascal}, and the assumption of independent and identical data distribution between training and testing samples. However, performance drops significantly when the model trained on training set (source domain) is directly applied to unseen test scenarios (target domain). In addition, densely annotating pixel-wise labels of many samples in target domain is time-consuming and uneconomical. 


To reduce the heavy demand for pixel-wise annotation, one way is to employ large amounts of easy-to-get simulation data which can be collected from game engines such as GTA5~\cite{richter2016playing} and SYNTHIA~\cite{ros2016synthia}. In addition, unsupervised domain adaptation (UDA) strategy, which aims at transferring knowledge from a synthetic label-rich source domain to a real-world label-scarce target domain, is required to bridge domain gap between synthetic and real-world domains. Impressive results have been achieved by UDA methods that extract domain-invariant representations via entropy minimization~\cite{pan2020unsupervised, vu2019dada}, generative modelling~\cite{hoffman2018cycada, gong2019dlow} and adversarial learning~\cite{vu2019advent, tsai2018learning}. However, domain shift cannot be completely eliminated by these methods due to weak supervision on target examples. There is still a big gap in performance compared with supervised methods. Another way in addressing the issue of heavy annotation is to annotate only a small set of images from target domain and make full use of plenty of unlabeled data with semi-supervised learning (SSL) techniques~\cite{french2019semi, ouali2020semi,feng2020semi,mittal2019semi}. Due to the shortage of labeled data in SSL setting, the obtained model has the risk of overfitting to the small amount of labeled data. How to effectively utilize available unlabeled and limited labeled data from different domains is the key in improving model's accuracy and generalization for pixel-wise prediction tasks.

Hence, a more practical task of semi-supervised domain adaptation (SSDA) is recently introduced by combining the small set of labeled target data images in SSL with the large amounts of labeled source domain data and unlabeled target domain data. In order to address the SSDA problem, one naive way is to equip UDA methods with additional supervision on the extra labeled target images (see Baseline in Table~\ref{tab:soto_cmp_gta}.). For example, Alleviating Semantic-level Shift (ASS) model~\cite{wang2020alleviating} is proposed for better promoting the distribution consistency of features by using adversarial learning on outputs from two labeled domains. However, these methods cannot fully explore rich information within available labeled and unlabeled data in two domains. 

Semantic segmentation is a dense pixel-wise prediction task, and classification of one pixel depends not only on its own value but also on its neighbourhood's context. We focus on how to effectively utilize labeled data in different domains to extract domain-invariant representations in both region-level and sample-level. The proposed framework consists of three steps. First, two kinds of data mixing methods are proposed to reduce domain gap in both region-level and sample-level. The region-level data mixing is achieved by applying two masks to labeled images from two domains and combining the two masked regions, which encourages a model to extract domain-invariant features about semantic structure from partial view. On the other hand, the image-level data mixing directly mixes labeled images from two domains from holistic view. Such two mixing ways help train two complementary teacher models that work on both two kinds of data distribution. In the second step, we employ knowledge distillation technique to extract ``dark knowledge'' from these two complementary teachers, which works as guidance in the learning process of a student model for target domain. By integrating knowledge from two views and making full use of unlabeled data, the student model of the same network architecture can give even better performance than any of its teachers. Once a good student model for target domain is obtained, pseudo labels could be generated with self-training strategy to expand the set of labeled target domain data for iterative update. Compared with traditional self-training methods, which directly use pseudo labels to train a final model, we instead leverage these pseudo labels to obtain two stronger domain-mixed teachers, which also leads to stronger student by another round of knowledge distillation. Overall, in our framework, teachers and student are progressively growing, and we can obtain a final well-trained student model. 

Our contributions of this paper are three-fold:
\setlist{nolistsep}
\begin{itemize}[noitemsep]
	\setlength{\itemsep}{2pt}
	\setlength{\parsep}{0pt}
	\setlength{\parskip}{0pt}
	\item Two kinds of data mixing methods are proposed to train domain-mixed teachers across domains in both region-level and sample-level to alleviate data distribution mismatch between different domains.
	\item A stronger student model on target domain can be obtained by distilling knowledge from complementary domain-mixed teachers. It can be further strengthened by employing pseudo labels which are generated for unlabeled target data in a self-training manner.
	\item Extensive experiments demonstrate that the proposed method can achieve superior performance on two common synthetic-to-real semantic segmentation benchmarks with small amounts of labeled data.
\end{itemize}

\section{Related Works}

\begin{figure*}[ht]
	\begin{center}
		\includegraphics[width=0.9\linewidth]{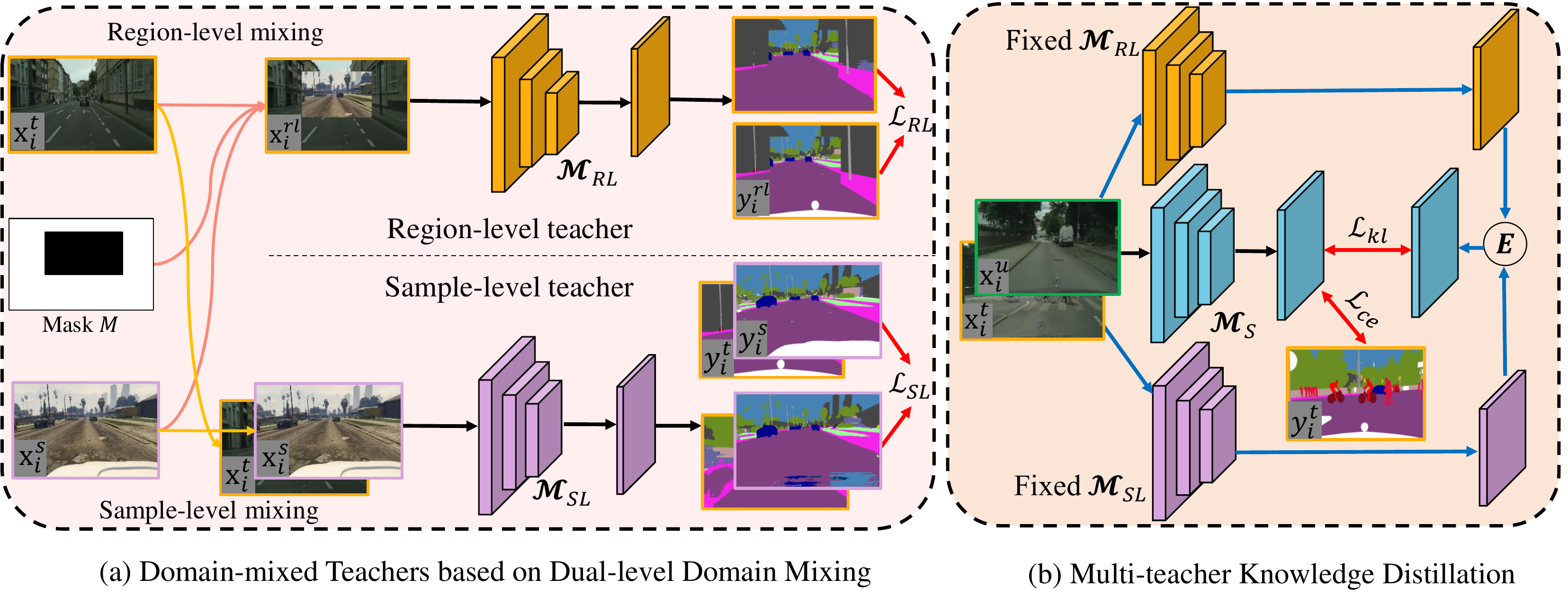}
	\end{center}
	\vspace{-10pt}
	\caption{The first two stages of the proposed SSDA framework, training of domain-mixing teachers and multi-teacher knowledge distillation. The domain-mixed teachers are trained based on the dual-level mixed data. Then these two domain-mixed teachers are used to train a good student. The student will generate pseudo labels for next round of teachers training. $\E$ means the ensemble operation of two domain-mixed teachers. The \textcolor{black}{black} arrows represent the training data flow, \textcolor{blue}{blue} arrows represent the data flow of inference, which do not require gradient backward. The \textcolor{red}{red} arrows represent the computation of losses.}
	\label{fig:framework}
	\vspace{-10pt}
\end{figure*}

\label{related}
\paragraph{Unsupervised domain adaptation for semantic segmentation.} Unsupervised domain adaptation (UDA) methods for semantic segmentation has been extensively studied to address domain discrepancy between photo-realistic synthetic dataset and unlabeled real dataset.
One mainstream approach is by adversarial learning~\cite{vu2019advent, tsai2018learning, chen2017no, chen2018road, hoffman2016fcns, saito2017adversarial, isobe2021multi}, which aims to employ a discriminator to measure the divergence across two domains.
Another approach to solving UDA problem is to utilize generative networks~\cite{sankaranarayanan2018learning, hoffman2018cycada, zhu2018penalizing} to generate target-style images by applying style transfer technique on annotated source image. 
Some methods based on self-training~\cite{kim2020learning, zou2018unsupervised, li2019bidirectional, he2021multi} have been employed to generate pseudo labels of unlabeled data and use them to fine-tune the model.~\cite{kim2020learning} firstly generate different stylized annotated images to learn texture-invariant representation and then use self-training to generate pseudo labels of unlabeled data to fine-tune the model on target domain.

Although impressive results have been achieved in UDA for semantic segmentation, the domain gap cannot be fully alleviated due to the lack of strong supervision in the target domain, and there is still an observed performance gap compared with their supervised counterparts.

\noindent\textbf{Semi-supervised learning for semantic segmentation.}
One way to reduce the heavy demand for manual pixel-wise labeling is to only label a small amount of data from target distribution and adopt semi-supervised learning (SSL) strategy to learn a great generalized model among ample unlabeled and limited labeled data.
Numerous methods have since been developed to improve model generalization~\cite{ouali2020semi,kervadec2019curriculum,feng2020semi,mittal2019semi, hung2018adversarial, chen2019multi, grandvalet2005semi}. Consistency regularization is one of the most popular methods and the key idea is to encourage the network to give consistent predictions for perturbed unlabeled inputs.
One most related work is~\cite{french2019semi}, which enforces a consistency between mixed output of teacher network and the prediction of student over the mixed inputs by a region-level data augmentation CutMix~\cite{yun2019cutmix} with a teacher-student architecture~\cite{tarvainen2017mean}. Our method also shares similar philosophy as theirs, however, we propose to train two domain-mixed teachers with two kinds of domain-mixing methods to fully exploit two sets of data from two different domains.

\noindent\textbf{Semi-supervised domain adaptation.} Also aims to reduce the data distribution mismatch, compared with UDA, semi-supervised domain adaptation (SSDA) bridges domain discrepancy via introducing partially labeled target samples. Recently, a few methods have been proposed based on deep learning~\cite{yang2020mico, qin2020opposite, kim2020attract, rukhovich2019mixmatch} for image classification.
~\cite{yang2020mico} decomposes SSDA into two sub-problems: UDA and SSL, and employ co-training~\cite{chen2011co} to exchange the expertise between two classifiers,  which are trained on MixUp-ed~\cite{zhang2017mixup} data between labeled and unlabeled data of each view.


Due to the complex densely pixel-wise prediction and no explicit decision boundaries between examples in semantic segmentation, SSDA methods based on discriminative class boundaries for image classification cannot be directly applied to semantic segmentation. Just one previous work have been developed to study SSDA for semantic segmentation. Wang \etal~\cite{wang2020alleviating} propose Alleviating Semantic-level Shift (ASS) framework to realize feature alignment across domain from global and semantic level. ASS introduces an extra semantic-level adaptation module through adversarial training on the corresponding outputs of source and target labeled inputs besides the additional supervision on extra small amount of labeled target data upon the classical AdaptSeg framework~\cite{tsai2018learning}. However, the naive supervision of labeled target samples cannot fully take advantage of labeled two domains, and the adversarial loss makes training unstable due to the weak supervision. To solve this issue, we propose a novel iterative framework based on dual-level domain mixing methods without any adversarial training.

\section{Method}
\label{method}
\subsection{Problem Statement}
In the setting of semi-supervised domain adaptation (SSDA), we are provided with a small set of labeled target domain images upon the large amounts of labeled source and unlabeled target domain images. Let $D_S = \{(\mathbf{x}_i^s, y_i^s)\}_{i=1}^{N_S}$ represents the $N_S$ labeled source domain samples, and $D_T = \{(\mathbf{x}_i^t, y_i^t)\}_{i=1}^{N_T}$ represents the $N_T$ labeled target domain samples, and $D_U = \{\mathbf{x}_i^u\}_{i=1}^{N_U}$ represents the $N_U$ unlabeled target domain samples. With the SSDA setting, we aim at developing a way to efficiently utilize the available $D_S$, $D_T$ and $D_U$ and obtain a segmentation model which has great performance on unseen test data sampled from target data distribution.

\subsection{Domain-mixed Teachers}
Performance degradation comes from inconsistent data distribution in different domains. We propose two data mixing methods for domain adaptation, one is region-level data mixing and the other is sample-level data mixing, to reduce the data distribution gap from two views. As we all know, data with labeled ground truth provides much information for training one model in deep learning-based methods. In SSDA, two types of labeled data, \ie, $D_S$, $D_T$, are provided. Our region-level and sample-level data mixing methods are implemented on these two kinds of labeled data, and two domain-mixed teacher models can be trained on the mixed data. Because of different views of data mixing, these two domain-mixed teachers are complementary.
\\
\noindent\textbf{Region-level data mixing.}
Semantic segmentation is a dense pixel-wise prediction task, and the classification of one pixel depends not only on its own value but also on its regional neighbourhood's context. Thus, if one image contains both source domain and target domain content, the model can learn domain-invariant representation because different regions with different feature distribution can be seen at the same time during model training.

Inspired by CutMix~\cite{yun2019cutmix} where patches from an image are cut and pasted to another one to augment data for improving model's generalization ability, here we propose to conduct region-level data mixing on set $D_S$ and $D_T$ to reduce domain gap.
Given two labeled images $\{\mathbf{x}^t, y^t\}$, $\{\mathbf{x}^s, y^s\}$, the region-level mixing operation can be described as below.
\begin{equation}
\begin{aligned}
&\mathbf{x}^{rl} = M \odot \mathbf{x}^t + (1-M) \odot \mathbf{x}^s, \\
&y^{rl} = M \odot y^t + (1-M) \odot y^s,
\label{eq:cutmix}
\end{aligned}
\end{equation}
where $M$ denotes a binary mask indicating where the region needs to fusion, and $\odot$ is element-wise multiplication. As shown in Fig.~\ref{fig:framework}, the mixed image $\mathbf{x}^{rl}$ contains both contents of $\mathbf{x}^s$ and $\mathbf{x}^t$, and the corresponding mixed labels $y^{rl}$ are obtained for each pixel according to which domain the region containing that pixel comes from. In detail, a rectangular region is cropped from $\mathbf{x}^s$ according to randomly chosen coordinates, and then pasted on the same location of $\mathbf{x}^t$. 
The region-level data mixing is able to produce intermediate samples between different domains, which works
as a bridge, filling in the gap between domains. This helps explore essential semantic contexts across different domains from partial view. Additionally, this operation can destroy the inherent structure of the original target picture, and regularize the training process of region-level teacher.
Once the mixed images and their labels are ready, we can train a semantic segmentation model through supervised training on the mixed data. The training objective function can be written as follows.
\begin{equation}
\begin{aligned}
\L_{RL} = \L_{ce}(\M_{RL}(\mathbf{x}^{rl}), y^{rl}),
\label{eq:CE_region}
\end{aligned}
\end{equation}
where $\M_{RL}$ represents teacher model trained on region-level mixed data, $\L_{ce}$ denotes the cross entropy loss.
\\
\noindent\textbf{Sample-level data mixing.}
Sample-level data mixing aims to mix the data from different domains from holistic view. The source and target examples are sampled from inconsistent distribution with a big gap. We find that direct mixing of these data can already help reduce the gap between different domains to some extent. There are two advantages with sample-level data mixing method. On the one hand, the introduction of large amounts of source images alleviates the model overfitting to the small amount of target images. On the other hand, the sample-level mixing helps explore intermediate decision boundary between different domains from holistic view. In our experiments, we randomly sample two examples from source set $D_S$ and target set $D_T$, then directly feed both of them into model during one iteration. Given two images from $D_S$ and $D_T$, the training objective function of sample-level teacher is defined as follows.
\begin{equation}
\begin{aligned}
\L_{SL} = \L_{ce}(\M_{SL}(\mathbf{x}^{s}), y^{s}) + \L_{ce}(\M_{SL}(\mathbf{x}^{t}), y^{t}),
\label{eq:CE_sample}
\end{aligned}
\end{equation}
where $\M_{SL}$ represents teacher model trained on sample-level mixed data.
\subsection{Multi-teacher Knowledge Distillation}
After obtaining two pre-trained domain-mixed teachers, we employ knowledge distillation (KD), a technique to distilling knowledge by minimizing the KL-divergence between outputs of these two models. Here we adapt it to extract ``dark knowledge'' from these two complementary teachers. The pipeline of multi-teacher KD is shown in Fig.~\ref{fig:framework} (b), including two pre-trained domain-mixed teachers and one student with the same network architecture as teacher. The outputs of two teachers are ensembled as a stronger guidance to supervise the training of the student model on unlabeled target data. Besides, the student model is also supervised by the labels on the small amount of labeled target data. The objective function of learning student $\M_S$ is defined as below.
\begin{equation}
\label{eq:KL}
\begin{aligned}
\L_{S} = &\lambda_{kl}\L_{kl}(\mathit{E}(\M_{RL}(\mathbf{x}^{u}), \M_{SL}(\mathbf{x}^{u})), \M_S(\mathbf{x}^{u})) \\
&+ \lambda_{ce}\L_{ce}(\M_S(\mathbf{x}^t), y^t),
\end{aligned}
\end{equation}
where $\lambda_{kl}$ and $\lambda_{ce}$ are the weights of KL-divergence loss and cross entropy loss respectively, $\E$ denotes the ensemble operation of two models. In experiments, the ensemble operation is implemented by averaging the outputs of two complementary teachers.

By integrating knowledge from two views and making full use of unlabeled data, we can obtain one student with even superior performance than any one of its teachers.
\begin{algorithm}[t]  
	\caption{Training process of our proposed framework.}  
	\label{alg:Framwork}  
	\begin{algorithmic}[1]  
		\Require  
		labeled source dataset $D_S = \{(\mathbf{x}_i^s, y_i^s)\}_{i=1}^{N_S}$,
		unlabeled target dataset $D_U = \{\mathbf{x}_i^u\}_{i=1}^{N_U}$,
		labeled target dataset $D_T = \{(\mathbf{x}_i^t, y_i^t)\}_{i=1}^{N_T}$, initialized weights of teachers model $\M_{RL}^0$, $\M_{SL}^0$ and student model $\M_S^0$, iterative rounds $R$.
		\Ensure
		\For{$r$ $\leftarrow$ 1 to $R$}
		\State Dual-level domain mixing 
		\State Optimize $\M_{RL}^r$ and $\M_{SL}^r$ by Eq. (\ref{eq:CE_region}) and (\ref{eq:CE_sample}) \Comment{Training two teachers}
		\State Optimize $\M_S^r$ by Eq. (\ref{eq:KL}) \Comment{Training student model}
		\State Generate pseudo labels $\hat{y}_i^u$ following~\cite{li2019bidirectional} by $\M_S^r$
		\State Update $D_U = \{\mathbf{x}_i^u, \hat{y}_i^u\}_{i=1}^{N_U}$ to labeled target dataset $D_T$
		\EndFor \\
		\Return student model $\M_S^R$
	\end{algorithmic}

\end{algorithm}
\subsection{Progressive Improving Scheme}
Normally, a teacher network usually has stronger ability than student network. However, here a good student model is obtained by distilling knowledge from the ensembled outputs of two complementary domain-mixed models on large amount of unlabeled data. We focus on how to use a student to further improve teachers' performance for next step. 

Recently, self-training as a simple but effective technique to address the scarceness of labeled training data, and are widely applied in SSL and UDA for image classification task. In our task, the teachers we obtain are trained based just on labeled source data and a small amount of labeled target data. Motivated by the success of self-training, we believe the teachers can be further improved with this strategy. 
In detail, following~\cite{li2019bidirectional}, pseudo labels of $D_U$ are generated via the learned student model to update the labeled set of images in dataset of $D_T$ for next round training of domain-mixed teachers. 
Once stronger domain-mixed teachers are obtained, a stronger student can be obtained by another round of multi-teacher KD.

Overall, the whole training process of our framework goes iteratively. Both the domain-mixed teachers and student are progressively growing, \ie, they can help the learning of each other through knowledge distillation and self-training strategies. We summarize our proposed algorithm in Algorithm~\ref{alg:Framwork}.
\begin{figure*}[t]
	\begin{center}
		\includegraphics[width=0.85\linewidth]{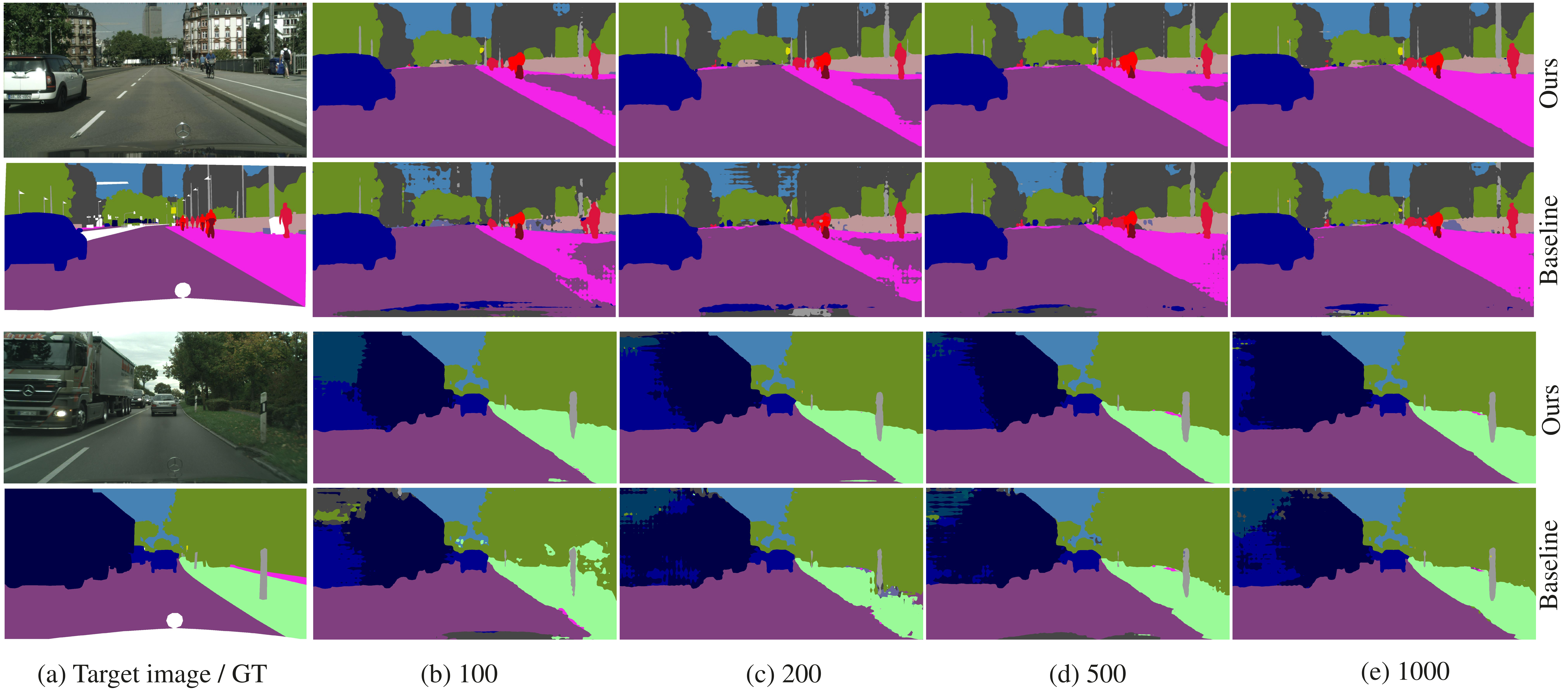}
	\end{center}
\vspace{-10pt}
	\caption{Qualitative results of our method and baseline method on different numbers of labeled target images on GTA5 to Cityscapes. (a) target images and corresponding ground truth (GT), (b)-(e) segmentation results of different numbers of labeled target images.}
	\label{fig:diff_num}
	\vspace{-10pt}
\end{figure*}
\section{Experiments}
\label{experiment}
\subsection{Experimental Setup}
Following the setting of unsupervised domain adaptation methods in semantic segmentation, we also conduct extensive experiments and report the mean intersection-over-union (mIoU) score on two commonly used synthetic-to-real benchmarks, which are GTA5~\cite{richter2016playing} and SYNTHIA~\cite{ros2016synthia} to Cityscapes~\cite{cordts2016cityscapes} respectively.

\textbf{Cityscapes} is an autonomous driving dataset captured from 50 cities in real world. It contains densely annotated 2,975 and 500 images with a fixed resolution of 2048$\times$1024 for training and validation respectively. All images are manually labeled by 19 semantic categories. For SSDA setting, we randomly select different numbers of images, such as (100, 200, 500, 1000), from the whole training set to demonstrate the effectiveness of our method across different settings. The validation set is used to evaluate the performance of our method.

\textbf{GTA5} is a synthetic dataset in which the images are collected from game video and the corresponding semantic labels are automatically generated by computer graphics techniques. It includes 24,966 synthesized images with pixel-wise labels of 33 classes. In experiments, we consider the 19 common classes with Cityscapes dataset to train our models. 

\textbf{SYNTHIA} is also a synthetic dataset and we use SYNTHIA-RAND-CITYSCAPES as another labeled source domain, which contains 9,400 fully annotated synthetic images with resolution of 1280$\times$960. It has 16 common categories with Cityscapes dataset. We train our models with the common classes and report the 13-class mIoU on validation set.
\begin{table}[t]
	\centering
	\caption{Semantic segmentation performance comparison with the state-of-the-art UDA, SSL and SSDA methods on GTA5$\rightarrow$Cityscapes. 19-class mIoU (\%) score are reported on Cityscapes validation set across 0, 100, 200, 500, 1000, 2975 numbers of labeled target images. ``$^*$'' denotes our reimplementation on corresponding numbers of labeled Cityscapes images. GTA5 images are not introduced for implementing SSL methods. Best results are \textbf{highlighted}.}
	\label{tab:soto_cmp_gta}
	\vspace{1mm}
	\small
	\setlength{\tabcolsep}{0.78mm}
	\begin{tabular}{l|l|cccccc}
		\hline
		\hline
		\multirow{2}{*}{Type} & \multirow{2}{*}{Methods} & \multicolumn{6}{c}{Labeled target images} \\
		&& 0 & 100 & 200 & 500 & 1000 & 2975\\
		\hline
		\multirow{4}{*}{UDA} & AdaptSeg~\cite{tsai2018learning} & 42.4 & - & - & - & - & -\\
		& Advent~\cite{vu2019advent} & 44.8 & - & - & - & - & -\\
		& LTI~\cite{kim2020learning} & 50.2 & - & - & - & - & -\\
		& PIT~\cite{lv2020cross} & \textbf{50.6} & - & - & - & - & -\\
		\hline
		Supervised & DeeplabV2 & - & 41.9 & 47.7 & 55.5 & 58.6 & 65.3\\
		\hline
		\multirow{2}{*}{SSL} & CutMix$^*$~\cite{french2019semi} & - & 50.8 & 54.8 & 61.7 & 64.8 & -\\
		& DST-CBC$^*$~\cite{feng2020semi} & - & 48.7 & 54.1 & 60.6 & 63.2 & -\\
		\hline
		\multirow{3}{*}{SSDA} & Baseline & - & 52.6 & 53.6 & 58.4 & 61.6 & 66.6 \\
		& MME$^*$~\cite{saito2019semi} & - & 52.6 & 54.4 & 57.6 & 61.0 &  64.2 \\
		& ASS~\cite{wang2020alleviating} & - & 54.2 & 56.0 & 60.2 & 64.5 & 69.1 \\
		& Ours & - & \textbf{61.2} & \textbf{60.5} & \textbf{64.3} & \textbf{66.6} & \textbf{69.8}\\
		\hline
	\end{tabular}
\vspace{-10pt}
\end{table}
\subsection{Implementation Details}
For all the following experiments, similar to~\cite{vu2019advent}, a DeeplabV2~\cite{chen2017deeplab} model, which contains Atrous Spatial Pyramid Pooling (ASPP) module to extract multi-scale representations and utilizes a pre-trained ResNet-101~\cite{he2016deep} on ImageNet as backbone, are employed as our semantic segmentation architecture. To train our proposed framework, we implement it using Pytorch deep learning toolbox. All the experiments are conducted on a single Tesla V100 GPU with 32GB memory to accelerate computing.

An optional operation before training the model is to apply a simple image translation method to source domain images to reduce the visual difference between source and target domain. Here images are converted into LAB color space and are matched to the statistics of target domain. Image translation is applied at the beginning in most experiments except as otherwise noted.
Then sample-level and region-level data mixing are conducted on labeled source data with target style and the target data. We then train domain-mixed teachers with cross entropy loss on supervised data. Student model is obtained on both labeled and unlabeled target data with cross entropy and KL loss. The weight $\lambda_{kl}$, $\lambda_{ce}$ in Eq.~\ref{eq:KL} are set to 0.5 and 1 respectively. For self-training, the portion of selected pseudo labels and the confidence threshold are separately, similar to~\cite{li2019bidirectional}, set to 0.5 and 0.9. Iterative rounds $R$ are set to 4 and 3 for GTA5$\rightarrow$Cityscapes and SYNTHIA$\rightarrow$Cityscapes respectively. All the models are trained by the Stochastic Gradient Descent (SGD) optimizer with initial learning rate 2.5$\times$$10^{-4}$, the momentum 0.9 and weight decay $10^{-4}$ as mentioned in~\cite{vu2019advent}. The learning rate is decreased with the polynomial annealing procedure with power of 0.9.
\begin{table}[t]
	\centering
	\caption{Semantic segmentation performance comparison with the state-of-the-art UDA, SSL and SSDA methods on SYNTHIA$\rightarrow$Cityscapes. Here we train the DeeplabV2 model with 16 classes and report 13-class mIoU (\%) score following the previous works on UDA. Other settings are kept same as in Table~\ref{tab:soto_cmp_gta}. Best results are \textbf{highlighted}.}
	\label{tab:soto_cmp_synthia}
	\vspace{1mm}
	\small
	\setlength{\tabcolsep}{0.78mm}
	\begin{tabular}{l|l|cccccc}
		\hline
		\hline
		\multirow{2}{*}{Type} & \multirow{2}{*}{Methods} & \multicolumn{6}{c}{Labeled target images} \\
		&& 0 & 100 & 200 & 500 & 1000 & 2975 \\
		\hline
		\multirow{4}{*}{UDA} & AdaptSeg~\cite{tsai2018learning} & 46.7 & - & - & - & - & -\\
		& Advent~\cite{vu2019advent} & 48.0 & - & - & - & - & -\\
		& LTI~\cite{kim2020learning} & 49.3 & - & - & - & - & -\\
		& PIT~\cite{lv2020cross} & \textbf{51.8} & - & - & - & - & -\\
		\hline
		Supervised & DeeplabV2 & - & 53.0 & 58.9 & 61.0 & 67.5 & 72.2\\
		\hline
		\multirow{2}{*}{SSL} & CutMix$^*$~\cite{french2019semi} & - & 61.3 & 66.7 & 71.1 & 73.0 & -\\
		& DST-CBC$^*$~\cite{feng2020semi} & - & 59.7 & 64.3 & 68.9 & 70.5 & - \\
		\hline
		\multirow{3}{*}{SSDA} & Baseline & - & 58.5 & 61.9 & 64.4 & 67.6 & 73.1\\
		& MME$^*$~\cite{saito2019semi} & - & 59.6 & 63.2 & 66.7 & 68.9 &  72.7 \\
		& ASS~\cite{wang2020alleviating} & - & 62.1 & 64.8 & 69.8 & 73.0 & 77.1\\
		& Ours & - & \textbf{68.4} & \textbf{69.8} & \textbf{71.7} & \textbf{74.2} & \textbf{77.2}\\
		\hline
	\end{tabular}
\end{table}
\begin{table*}[ht]
	\centering
	\caption{Performance comparison of ensembled models from single view and inconsistent views in terms of per-class IoUs and mIoU (\%). The $\M_{RL}^{1'}$ means the repeat run of the first round of $\M_{RL}^{1}$. Best results are \textbf{highlighted}.}
	\label{tab:view_cmp}
	\vspace{1mm}
	\small
	\renewcommand\arraystretch{1.1}
	\setlength{\tabcolsep}{0.78mm}
	\begin{tabular}{l|ccccccccccccccccccc|c}
		\hline
		\hline
		\multicolumn{21}{c}{GTA5$\rightarrow$Cityscapes}\\
		\hline
		Model & \rotatebox{90}{road} & \rotatebox{90}{sidewalk} & \rotatebox{90}{building} & \rotatebox{90}{wall} & \rotatebox{90}{fence} & \rotatebox{90}{pole} & \rotatebox{90}{light} & \rotatebox{90}{sign} & \rotatebox{90}{veg.} & \rotatebox{90}{terrain} & \rotatebox{90}{sky} & \rotatebox{90}{person} & \rotatebox{90}{rider} & \rotatebox{90}{car} & \rotatebox{90}{truck} & \rotatebox{90}{bus} & \rotatebox{90}{train} & \rotatebox{90}{mbike} & \rotatebox{90}{bike} & mIoU \\
		\hline
		$\M_{RL}^{1}$ & 95.0 & 66.6 & 85.4 & 19.7 &	20.1 & 38.7 & 37.3 & 50.6 &	87.5 & 46.6 & 89.5 & 65.2 &	33.9 & 89.1 & 46.8 & 37.6 &	13.1 & 35.1 & 59.6 & 53.5 \\
		$\M_{RL}^{1'}$ & 95.5 & 67.6 & 85.6 & 30.5 & 19.6 & 36.9 & 35.0 & 49.6 & 87.6 & 46.4 & \textbf{90.0} & 64.8 & 29.5 & 88.3 & 37.7 & 39.1 & \textbf{15.0} & 37.7 & 58.8 & 53.4 \\
		\hline
		$\M_{SL}^{1}$ & 93.8 & 59.2 & 85.4 & 33.9 & 29.0 & 37.5	& \textbf{42.3} & 45.3 &	86.8 & 44.4 & 86.0 & 63.8 &	37.6 & 87.7 & 45.9 & 49.5 &	0.1 & 39.0 & 56.2 & 53.9 \\
		$\M_{SL}^{1'}$ & 94.1 & 61.2 & 85.6 & 32.5 & 31.2 & 38.2 & 38.7 & 44.6 & 86.4 & 46.3 & 86.6 & 64.1 & 38.3 & 88.3 & 42.9 & 48.5 & 1.5 & 35.9 & 54.7 & 53.7 \\
		\hline
		$\E(\M_{RL}^{1}, \M_{RL}^{1'})$ & \textbf{95.6} & \textbf{68.4} & 86.0 & 24.6 & 20.8 & 38.8 & 37.6 & \textbf{50.9} & 87.8 & 47.7 & 89.7 & 65.4 & 33.1 & 89.2 & 44.7 & 41.6 & 10.9 & 35.3 & 60.2 & 54.1 \\
		$\E(\M_{SL}^{1}, \M_{SL}^{1'})$ & 94.4 & 62.1 & 86.0 & 33.4 & \textbf{31.7} & 38.9 & 42.0 & 45.6 & 87.2 & 47.6 & 87.0 & 64.9 & 39.1 & 88.8 & 48.4 & 50.8 & 0.5 & 39.9 & 57.0 & 55.0 \\
		$\E(\M_{RL}^{1}, \M_{SL}^{1})$ & 95.5 & 67.4 & 86.0 & 30.1 & 26.3 & 39.6 & 41.7 & 50.0 & \textbf{88.0} & 49.3 & 89.1 & \textbf{66.7} & \textbf{40.0} & \textbf{90.0} & \textbf{53.2} & 49.6 & 0.7 & \textbf{43.3} & \textbf{61.3} & \textbf{56.2} \\
		$\E(\M_{RL}^{1}, \M_{SL}^{1'})$ & 95.5 & 68.0 & 86.1 & 30.8 & 28.5 & \textbf{40.2} & 39.2 & 49.4 & 87.9 & 50.0 & 89.1 & \textbf{66.7} & \textbf{40.0} & 89.8 & 50.0 & 48.0 & 2.8 & 42.6 & 59.9 & 56.0 \\
		$\E(\M_{RL}^{1'}, \M_{SL}^{1})$ & 95.5 & 67.8 & \textbf{86.4} & 33.0 & 26.9 & 39.3 & 42.1 & 49.5 & \textbf{88.0} & 50.2 & 89.1 & 66.3 & 39.1 & 89.8 & 50.6 & \textbf{51.2} & 0.2 & 42.0 & 60.4 & \textbf{56.2} \\
		$\E(\M_{RL}^{1'}, \M_{SL}^{1'})$ & 95.5 & 68.2 & \textbf{86.4} & \textbf{33.7} & 29.0 & 39.9 & 39.6 & 48.9 & \textbf{88.0} & \textbf{50.5} & 89.1 & 66.4 & 39.7 & 89.7 & 46.7 & 49.3 & 1.2 & 41.4 & 59.4 & 55.9 \\
		\hline
	\end{tabular}
\end{table*}
\begin{table*}[t]
	\centering
	\caption{The detailed results of domain-mixed teachers and student model during different rounds in the whole training process on GTA5$\rightarrow$Cityscapes. For 2975 labeled images, our framework is justly trained one round.}
	\label{tab:self-training}
	\vspace{1mm}
	\small
	\renewcommand\arraystretch{1.2}
	\setlength{\tabcolsep}{0.78mm}
	\begin{tabular}{l|cccc|cccc|cccc|cccc|cccc}
		\hline
		\hline
		Number & \multicolumn{4}{c}{100} &  \multicolumn{4}{c}{200} & \multicolumn{4}{c}{500} & \multicolumn{4}{c}{1000} & \multicolumn{4}{c}{2975}\\
		\hline
		Rounds $R$ & 1 & 2 & 3 & 4 & 1 & 2 & 3 & 4 & 1 & 2 & 3 & 4 & 1 & 2 & 3 & 4 & 1 & 2 & 3 & 4\\
		\hline
		$\M_{RL}^{r}$ & 53.5 & 59.6 & 59.9 & 60.4 & 56.6 & 59.2 & 59.8 & 59.1 & 61.7 & 62.7 & 63.1 & 63.6 & 65.4 & 65.3 & 65.4 & 64.7 & 68.2 & - & - & - \\
		$\M_{SL}^{r}$ & 53.9 & 57.8 & 59.7 & 59.3 & 54.4 & 57.3 & 58.9 & 58.5 & 58.4 & 61.0 & 61.1 & 61.4 & 61.7 & 63.1 & 63.7 & 63.3 & 65.8 & - & - & -\\
		$\M_S^r$ & 57.1 & 59.8 & 61.0 & 61.2 & 58.3 & 60.2 & 60.3 & 60.5 & 62.5 & 63.7 & 64.1 & 64.3 & 65.5 & 66.0 & 66.6 & 66.0 & 69.8 & - & - & -\\
		\hline
	\end{tabular}
\vspace{-10pt}
\end{table*}
\subsection{Performance Comparison}
Our proposed method is conducted on two common synthetic-to-real GTA5 to Cityscapes and SYNTHIA to Cityscapes benchmarks to demonstrate the effectiveness of proposed framework. The performance is compared with the baseline method and existing state-of-the-art methods on UDA, SSL and SSDA settings. More extensive experiments can be seen in supplementary materials.
\\
\noindent\textbf{Baseline.} SSDA aims to alleviate the domain shift problem by introducing extra a small amount of labeled target data compared with UDA setting. As mentioned in Section~\ref{intro}, one naive way to address SSDA problem is by adding additional supervision upon UDA methods. Therefore, here we employ the classical UDA method named AdaptSeg~\cite{tsai2018learning}, one multi-level adaptation method by adversarial learning on multi-level outputs, with extra supervised cross entropy loss on limited labeled target images as our baseline model.
\\
\noindent\textbf{GTA5 to Cityscapes.} The performance comparisons with several state-of-the-art methods on GTA5 to Cityscapes are shown in Table~\ref{tab:soto_cmp_gta}. In experiment, iterative round $R$ is set to 4. After the iterative training, our method achieves the best performance on different ratios of labeled target domain images compared with existing methods in UDA, SSL and SSDA settings. Compared with UDA methods such as AdaptSeg, Advent~\cite{vu2019advent}, LTI~\cite{kim2020learning}, and PIT~\cite{lv2020cross}, our method can obtain above 10\% performance improvement by labeling just 100 target images and significantly reduce the performance gap compared with the oracle model. Particularly, our method outperforms the SSL methods CutMix~\cite{french2019semi} and DST-CBC~\cite{feng2020semi}, which use related CutMix and self-training techniques respectively, by a large performance gain. ASS~\cite{wang2020alleviating}, to be our known, which is the first work on SSDA for semantic segmentation, employs additional semantic-level adaptation on the outputs of both labeled source and target images to alleviate semantic-level shift except the additional supervision. We modify MME~\cite{saito2019semi}, which is used to address image classification in SSDA setting, for semantic segmentation task, and obtain inferior results. We think the reason is that SSDA methods for classification without taking into account the semantic contexts in an image and cannot be directly applied to segmentation task. The proposed approach obtains superior results on all ratios of labeled data. The reason is that the supervision of adversarial learning is weak and we can fully take advantage of available labeled data to reduce domain gap by dual-level data mixing. In addition, our method also performs well on fully 2975 images with the performance of 69.8\%.

In Fig.~\ref{fig:diff_num}, we further display some qualitative segmentation results of both our method and baseline method on 100, 200, 500 and 1000 labeled target images. Overall, our method achieves more complete segmentation results than baseline model in the same ratio of labeled images. As the number of labeled images increases, more refined segmentation results we can obtain by our proposed approach.
\\
\noindent\textbf{SYNTHIA to Cityscapes.} In order to further measure the performance of our approach, we also compare the results with several state-of-the-art methods on the SYNTHIA to Cityscapes. Since there are only 16 common categories between the SYNTHIA and Cityscapes, we just train a segmentation model with the common categories. As shown in Table~\ref{tab:soto_cmp_synthia}, following previous UDA works~\cite{vu2019advent, tsai2018learning}, we also report 13-class mIoU score to compare with existing other methods. From the results, it is clear that our method outperforms the UDA, SSL and SSDA methods with a large performance gain. And the similar discussions we can draw as in ``GTA5 to Cityscapes''. 

\begin{figure*}[t]
	\begin{center}
		\includegraphics[width=0.87\linewidth]{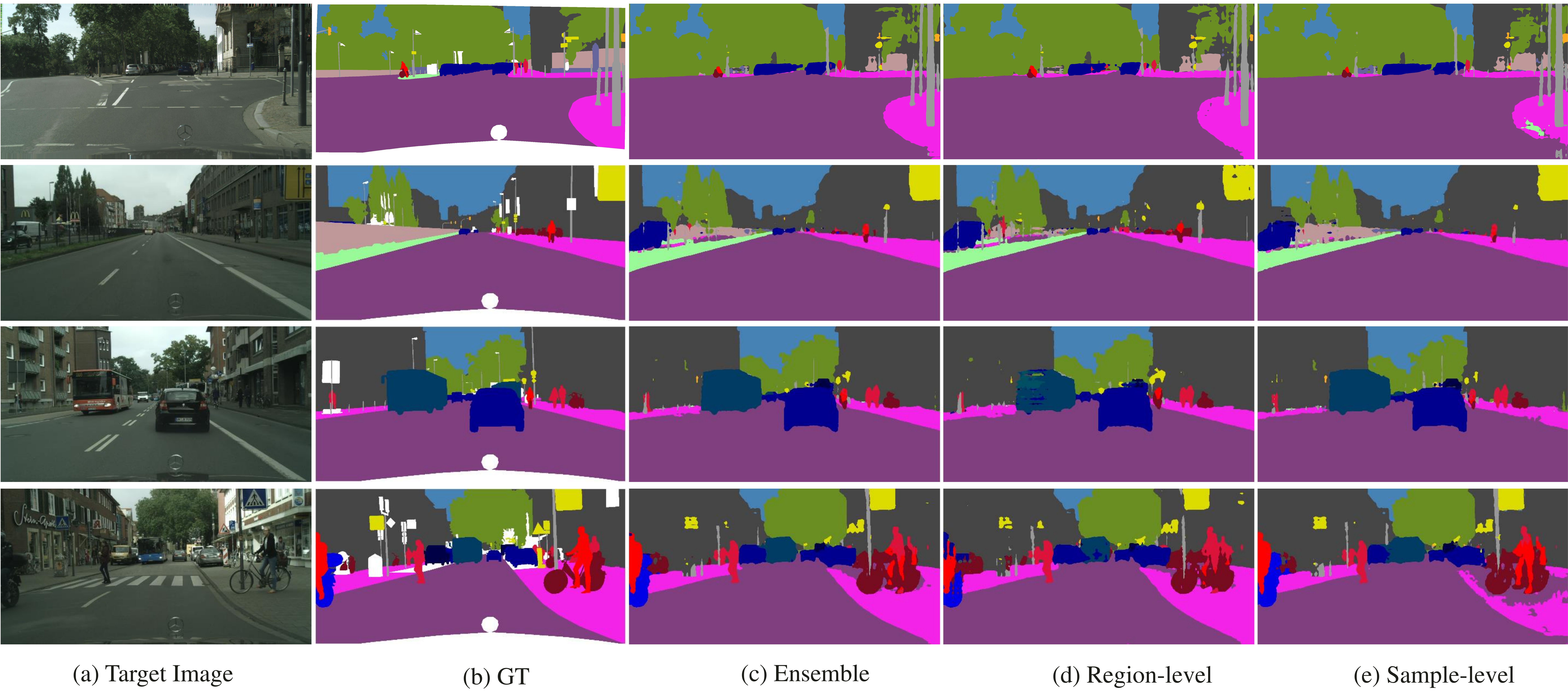}
	\end{center}
\vspace{-10pt}
	\caption{Qualitative results of ensembled models from region-level and sample-level views on GTA5 to Cityscapes. (a) target images, (b) ground truth, (c) segmentation results of model ensemble of different views, (d) results of model trained on region-level mixed data, (e) results of model trained on sample-level mixed data.}
	\label{fig:diff_views}
	\vspace{-10pt}
\end{figure*}

\section{Ablation Study}
\subsection{Complementarity}
To examine the complementarity of models trained from different views, we select 100 labeled target images and train the domain-mixed teachers twice separately from sample-level and region-level mixed data. Then ensemble of different models including two region-level teachers and two sample-level teachers, are conducted and results are shown in Table~\ref{tab:view_cmp}. Overall, from Table~\ref{tab:view_cmp}, we can draw a conclusion that model ensemble is effective for improving the performance, and the ensembled models from dual-level data mixing views can achieve better results than that from single-level data mixing view. In detail, region-level teachers perform better in categories that can be predicted without strictly relying on the structural information, such as road, sidewalk, vegetation and sky. However, they have poor predictions on the fence, light and bus classes whose shape is distinctive. We explain that the region-level data mixing operation could destroy the structure of these classes. Although model ensemble from one single view can realize impressive results on its own advantageous categories, the categories with poor performance are still poor. For example, the ensembled model from two region-level models achieves the best IoU score on road and sidewalk classes, and the worst results on rider and fence classes. Such best-worst phenomenon also occurs in ensembled model of sample-level teachers, but on different categories compared with ones in region-level. So we can fuse the models with different complementary levels and achieve a good result in all categories.

We also visualize some segmentation results of ensemble of different-level models in Fig.~\ref{fig:diff_views}. From Fig.~\ref{fig:diff_views}, the pixels what is wrongly classified in one view will be corrected in another view.
\subsection{Number of Iterative Rounds}
We discuss our results reported in Table~\ref{tab:soto_cmp_gta} during different rounds in the whole training  process on GTA5 to Cityscapes, and the detailed results of three models, domain-mixed teachers and student, are shown in Table~\ref{tab:self-training}. All three models can be improved with obvious performance gain compared with first round in training process. During different rounds, the student model will outperform both of two teachers, and the stronger student will correct the learning of teachers through generating more accurate pseudo labels, thus the teachers and student are progressive growing. This demonstrates the effectiveness of our proposed iteratively framework. We notice that the best models are achieved in different rounds on different numbers of labeled images.
\begin{table}[t]
	\centering
	\caption{Results of two domain-mixed teachers and student in the first round of our framework whether using style transfer or not across different number of labeled target images.}
	\label{tab:trans_cmp}
	\vspace{1mm}
	\small
	\setlength{\tabcolsep}{0.78mm}
	\begin{tabular}{l|c|ccccc}
		\hline
		\hline
		Model & Use-trans & 100 & 200 & 500 & 1000 & 2975 \\
		\hline
		\multirow{2}{*}{$\M_{RL}^1$} & T & 53.5 & 56.6 & 61.7 & 65.4 & 68.2 \\
		& F & 52.7 & 55.6 & 62.1 &  65.2 & 67.9 \\
		\hline
		\multirow{2}{*}{$\M_{SL}^1$} & T & 53.9 & 54.4 & 58.4 & 61.7 & 65.8 \\
		& F & 51.5 & 53.5 & 56.4 & 59.8 & 64.0 \\
		\hline
		\multirow{2}{*}{$\M_S^1$} & T & 57.1 & 58.3 & 62.5 & 65.5 & 69.8 \\
		& F & 55.8 & 57.0 & 62.1 & 65.0 & 68.9 \\
		\hline
	\end{tabular}
\vspace{-10pt}
\end{table}
\subsection{Image Translation}
In the above experiments, a simple image translation method in LAB color space is firstly taken to further reduce the visual difference between different domain images. Additionally, the experiments without style transfer are also conducted to demonstrate the effectiveness of our approach. We just compare the results of two-domain mixed teachers and student model in the first round of our framework. From Table~\ref{tab:trans_cmp}, we can draw the following three observations. First, the student model using style transfer achieves better performance than ones without it. Therefore, dual-level data mixing with style transfer can further reduce distribution mismatch across domains. Secondly, the teacher model trained on region-level mixed data becomes insensitive as the number of images increases. Superior performance without style transfer is obtained on 500 labeled target images than using style transfer. We argue that the reason is that region-level data mixing is relatively robust to whether style transfer is conducted in the one patch cropped from source image. Because of significant improvement in sample-level data mixing, we can also obtain better results with style transfer. Finally, our proposed framework can obtain better results than ASS even if without style transfer on 100, 200, 500, 1000 labeled images.

\section{Conclusion}
\label{conclusion}
In this paper, we propose a novel framework based on dual-level domain mixing to address semi-supervised domain adaptation problem. Two complementary domain-mixed teachers can be obtained based on proposed two kinds of data mixing methods in both region-level and sample-level. Then a stronger student model on target domain can be by distilling knowledge from these two domain-mixed teachers. Finally, pseudo labels can be generated by self-training manner for next round training of domain-mixed teachers. Extensive experiments demonstrate the proposed framework can fully take advantage of available data, and achieve superior performance on two commonly used synthetic-to-real benchmarks.

\clearpage
{\small
\bibliographystyle{ieee_fullname}
\bibliography{ref}
}
\appendix
\section{More Experiments}
\label{exp}
\subsection{Datasets}
\textbf{Synscapes}~\cite{wrenninge2018synscapes} is another photorealistic synthetic dataset for street scene parsing, which contains 25,000 RGB images with the resolution of 1440$\times$720. Synscapes is designed to be similar in structure and content to the real-world Cityscapes dataset~\cite{cordts2016cityscapes}, and it includes all 19 training classes for semantic segmentation in Cityscapes. To further verify the effectiveness of our method, we use the entire synthetic dataset as another source domain and consider 19 common categories to train our models on Synscapes to Cityscapes benchmark.
\subsection{Implementation Details}
\textbf{Architecture.} As the description in the main paper, we also utilize the DeepLabV2 with ResNet101 as the segmentation model. In detail, following~\cite{tsai2018learning}, we also adopt the multi-level adaptation architecture, which contains two additional ASPP modules on the last two convolutional layers, for fair comparison.

\begin{table}[H]
	\centering
	\caption{Semantic segmentation performance comparison with the state-of-the-art UDA, SSL and SSDA methods on Synscapes$\rightarrow$Cityscapes. 19-class mIoU (\%) score are reported on Cityscapes validation set across 0, 100, 200, 500, 1000, 2975 numbers of labeled target images. ``*'' denotes our reimplementation on corresponding numbers of labeled Cityscapes images. Synscapes images are not introduced for implementing SSL methods. Best results are \textbf{highlighted}.}
	\label{tab:soto_cmp_synscapes}
	\vspace{1mm}
	\small
	\setlength{\tabcolsep}{0.78mm}
	\begin{tabular}{l|l|cccccc}
		\hline
		\hline
		\multirow{2}{*}{Type} & \multirow{2}{*}{Methods} & \multicolumn{6}{c}{Labeled target images} \\
		&& 0 & 100 & 200 & 500 & 1000 & 2975\\
		\hline
		\multirow{2}{*}{UDA} & AdaptSeg$^*$~\cite{tsai2018learning} & 51.3 & - & - & - & - & -\\
		& Advent$^*$~\cite{vu2019advent} & \textbf{51.6} & - & - & - & - & -\\
		\hline
		Supervised & DeeplabV2 & - & 41.9 & 47.7 & 55.5 & 58.6 & 65.3\\
		\hline
		\multirow{2}{*}{SSL} & CutMix$^*$~\cite{french2019semi} & - & 50.8 & 54.8 & 61.7 & 64.8 & -\\
		& DST-CBC$^*$~\cite{feng2020semi} & - & 48.7 & 54.1 & 60.6 & 63.2 & -\\
		\hline
		\multirow{2}{*}{SSDA} & Baseline & - & 57.3 & 58.1 & 61.5 & 63.9 & 67.4 \\
		& MME$^*$~\cite{saito2019semi} & - & 56.6 & 57.1 & 60.6 & 63.1 &  67.9 \\
		& Ours & - & \textbf{62.0} & \textbf{62.5} & \textbf{65.1} & \textbf{68.2} & \textbf{71.0}\\
		\hline
	\end{tabular}
\end{table}
\begin{table}[t]
	\centering
	\caption{The results of students trained on single-teacher and multi-teacher knowledge distillation method. ``SL'' and ``RL'' denote the teacher model trained on sample-level mixed data and region-level mixed data, respectively. $\E$ means ensemble operation of two domain-mixed teachers. All the results are obtained at first round on GTA5$\rightarrow$Cityscapes.}
	\label{tab:single-multi}
	\vspace{1mm}
	\small
	\renewcommand\arraystretch{1.2}
	\setlength{\tabcolsep}{0.78mm}
	\begin{tabular}{l|c|ccccc}
		\hline
		\hline
		\multicolumn{2}{c}{Model} & 100 & 200 & 500 & 1000 & 2975\\
		\hline
		\multirow{2}{*}{SL} & $\M^1_{SL}$ & 53.9 & 54.4 & 58.4 & 61.7 & 65.8 \\
		& $\M^1_S$ & 55.9 & 56.2 & 61.5 & 64.5 & 68.1\\
		\hline
		\multirow{2}{*}{RL} & $\M^1_{RL}$ & 53.5 & 56.6 & 61.7 & 65.4 & 68.2 \\
		& $\M^1_S$ & 54.8 & 57.1 & 61.9 & 65.3 & 69.6 \\
		\hline
		\multirow{2}{*}{SL \& RL} & $\E(\M^1_{SL} , \M^1_{RL})$ & 56.2 & 57.5 & 62.3 & 65.8 & 69.1 \\
		& $\M^1_S$ & 57.1 & 58.3 & 62.6 & 65.5 & 69.8\\
		\hline
	\end{tabular}
\end{table}
\begin{table*}[h]
	\centering
	\caption{The detailed results of student model during different rounds through vanilla self-training and our proposed progressive improving scheme on GTA5$\rightarrow$Cityscapes.}
	\label{tab:self-training_ablation}
	\vspace{1mm}
	\small
	\renewcommand\arraystretch{1.2}
	\setlength{\tabcolsep}{0.78mm}
	\begin{tabular}{lc|cccc|cccc|cccc|cccc}
		\hline
		\hline
		\multicolumn{2}{c}{Number} & \multicolumn{4}{c}{100} &  \multicolumn{4}{c}{200} & \multicolumn{4}{c}{500} & \multicolumn{4}{c}{1000} \\
		\hline
		\multicolumn{2}{c|}{Rounds $R$} & 1 & 2 & 3 & 4 & 1 & 2 & 3 & 4 & 1 & 2 & 3 & 4 & 1 & 2 & 3 & 4\\
		\hline
		Vanilla & $\M_S^r$ & 57.1 & 56.7 & 55.1 & 53.3 & 58.3 & 57.9 & 57.6 & 56.5 & 62.5 & 60.8 & 59.3 & 58.4 & 65.5 & 62.8 & 61.6 & 60.3 \\
		Ours & $\M_S^r$ & 57.1 & 59.8 & 61.0 & 61.2 & 58.3 & 60.2 & 60.3 & 60.5 & 62.5 & 63.7 & 64.1 & 64.3 & 65.5 & 66.0 & 66.6 & 66.0 \\
		\hline
	\end{tabular}
\end{table*}
\textbf{Training Details.} During training, all the models are trained 250,000 iterations and early stopped at 120,000 iterations. Iterative rounds $R$ is set to 3 on Synscapes to Cityscapes. 
\subsection{Results on Synscapes to Cityscapes}
We show the results of our methods and several state-of-the-art methods on Synscapes to Cityscapes in Table~\ref{tab:soto_cmp_synscapes}. From Table~\ref{tab:soto_cmp_synscapes}, our approach obtains superior results on all ratios of labeled data compared with UDA and SSL methods on Synscapes to Cityscapes. Due to the similarity of style and content between these two datasets, significant performance improvement can be obtained by our method. It is noteworthy that our method achieves 71.0\% mIoU when using full data in target domain.

\section{More Ablation Studies}
\label{ablation}
\subsection{Single-teacher VS. Multi-teacher}
In our proposed framework, a good student can be obtained by distilling knowledge from multi domain-mixed teachers, \ie, teachers trained on sample-level and region-level mixed data. Here, we compare the results of students via different knowledge distillation from one single teacher and multi teachers. We just run first round of our iterative framework on GTA5~\cite{richter2016playing} to Cityscapes, and the results are shown in Table~\ref{tab:single-multi}. From Table~\ref{tab:single-multi}, one best student model is achieved by our multi-teacher knowledge distillation framework with a large performance gain at 100 and 200 labeled images. Thus more accurate pseudo labels generated by student model can promote the next round training of teachers. However, at 500, 1000 and 2975 labled images, the multi-teacher knowledge distillation has the weak advantage compared with single region-level teacher. We argue that compared with full labeled data, such a lot of labeled images will provide enough information especially for region-level data mixing to train a better teacher network. The rest of unlabeled target images cannot provide extra information for further improving the student model.
\subsection{Vanilla Self-training VS. Progressive Improving Scheme}
Self-training is proposed to address the scarceness of labeled training data and successfully used in UDA and SSL tasks. Vanilla self-training aims to generate pseudo labels of unlabeled data by one model and leverage them to retrain this model. We instead use the pseudo labels to train two stronger teachers. To further demonstrate the advantage of progressive improving scheme between domain-mixed teachers and student, we conduct the vanilla self-training method on the student model obtained at first round on GTA5 to Cityscapes. In experiments, the portion of selected pseudo labels and the confidence threshold are kept same and set to 0.5 and 0.9 respectively. Table~\ref{tab:self-training_ablation} shows the performance comparison between different self-training strategies. As the number of rounds increases, the performance of the student model obtained by vanilla self-training deceases. We explain that the initial student cannot be further improved through the pseudo labels generated by itself in our framework. The key of self-training is by generating pseudo labels of unlabeled data to further improve performance of model. However, the initial student model for self-training in our framework is obtained through the supervision of soft labels generated by ensemble of multi teachers on labeled and unlabeled target data, \ie, this supervsion of pseudo- or soft- label mechanism has been used in the process of obtaining the student model. In addition, soft label has the more robustness ability than pseudo label because wrong pixels usually existing in pseudo label. Thus the vanilla self-training will lead to the performance drop through the pseudo labels generated by itself. However, in the progressive improving scheme, we instead use the pseudo labels for training two domain-mixed teachers. Due to accurately labeled ground truths in source domain images, the wrong pixels in pseudo labels has less impact after two kinds of data mixing methods.
\end{document}